\documentclass[final]{cvpr}

\usepackage{times}
\usepackage{epsfig}
\usepackage{graphicx}
\usepackage{amsmath}
\usepackage{amssymb}
\usepackage{comment}
\usepackage{multirow}

\usepackage{color}

\usepackage[pagebackref=true,breaklinks=true,colorlinks,bookmarks=false]{hyperref}

\newcommand{\tablestyle}[2]{\setlength{\tabcolsep}{#1}\renewcommand{\arraystretch}{#2}\centering\small}
\newlength\savewidth\newcommand\shline{\noalign{\global\savewidth\arrayrulewidth
  \global\arrayrulewidth 1pt}\hline\noalign{\global\arrayrulewidth\savewidth}}

\def\cvprPaperID{****} 
\def\confYear{CVPR 2021}

\begin{document}

\title{Weakly-Supervised Temporal Action Localization Through \\Local-Global Background Modeling}

\author{
Xiang Wang$^{1,2}$
\quad Zhiwu Qing$^{1,2}$ 
\quad Ziyuan Huang$^{2}$ 
\quad Yutong Feng$^{2}$ 
\quad Shiwei Zhang$^{2*}$
\\
\quad Jianwen Jiang$^2$ 
\quad Mingqian Tang$^2$ 
\quad Yuanjie Shao$^1$
\quad Nong Sang$^{1*}$ 
\\
$^1$ Key Laboratory of Image Processing and Intelligent Control \\
School of Artificial Intelligence and Automation, Huazhong University of Science and Technology\\
$^2$Alibaba Group\\
{\tt\small \{wxiang, qzw, shaoyuanjie, nsang\}@hust.edu.cn}\\
{\tt\small \{pishi.hzy, fengyutong.fyt, zhangjin.zsw, jianwen.jjw, mingqian.tmq\}@alibaba-inc.com}
}

\maketitle

\let\thefootnote\relax\footnotetext{$*$ Corresponding authors.}
\let\thefootnote\relax\footnotetext{This work is supported by Alibaba Group through Alibaba Research Intern Program.}


\begin{abstract}
Weakly-Supervised Temporal Action Localization (WS-TAL) task aims to recognize and localize temporal starts and ends of action instances in an untrimmed video with only video-level label supervision.
Due to lack of negative samples of background category, it is difficult for the network to separate foreground and background, resulting in poor detection performance.
In this report, we present our 2021 HACS Challenge - Weakly-supervised Learning Track~\cite{HACS} solution that based on BaSNet~\cite{BaSNet} to address above problem. 
Specifically, we first adopt pre-trained CSN~\cite{csn}, Slowfast~\cite{slowfast}, TDN~\cite{TDN}, and ViViT~\cite{VIVIT} as feature extractors to get feature sequences.
Then our proposed Local-Global Background Modeling Network (LGBM-Net) is trained to localize instances by using only video-level labels based on Multi-Instance Learning (MIL). 
Finally, we ensemble multiple models to get the final detection results and reach \textcolor{blue}{\textbf{22.45\%}} mAP on the test set.

\end{abstract}

\section{Introduction}

Video understanding is an important area in computer vision, including many sub-research directions, such as Action Recognition~\cite{tsn,slowfast,huang2021self}, Temporal Action Detection~\cite{gtad,bmn,qing2021temporal,wang2021self}, Spatio-Temporal Action Detection~\cite{song2019tacnet,anetava2018}, \etc. 
In this report, we introduce our method for the temporal action detection task with only video-level supervision, termed weakly-supervised temporal action localization (WS-TAL).

Since the setting of weak supervision is more in line with real needs, WS-TAL has attracted more and more attention. Recently, several methods~\cite{luo2020weakly,islam2021hybrid,WTALC,STPN} were developed to localize instances in untrimmed videos using the video-level labels. However, though these methods have achieved significant performance, there is still a performance gab between fully-supervised methods~\cite{bmn,qing2021temporal,gtad}. We attribute this to that there are plenty of foreground-background confusions. It is challenging based on only video-level labels to separate action and background. To address this issue, we improve the mainstream approach BaSNet~\cite{BaSNet} and propose Local-Global Background Modeling Network (LGBM-Net), which integrates two-branch weight sharing local-global sub-net and a local-global attention module to suppress background and improve the discrimination of actions.

\section{Feature Extractor}
Following recent WS-TAL methods~\cite{WTALC,STPN}, given an untrimmed video $V$, we first divide it into multiple snippets based on a pre-defined sampling ratio, and then apply pre-trained networks to extract snippet-level features. Next, we briefly introduce the feature extraction networks we used in the competition.

\begin{figure*}[t!]
\centering
\centering{\includegraphics[width=.9\linewidth]{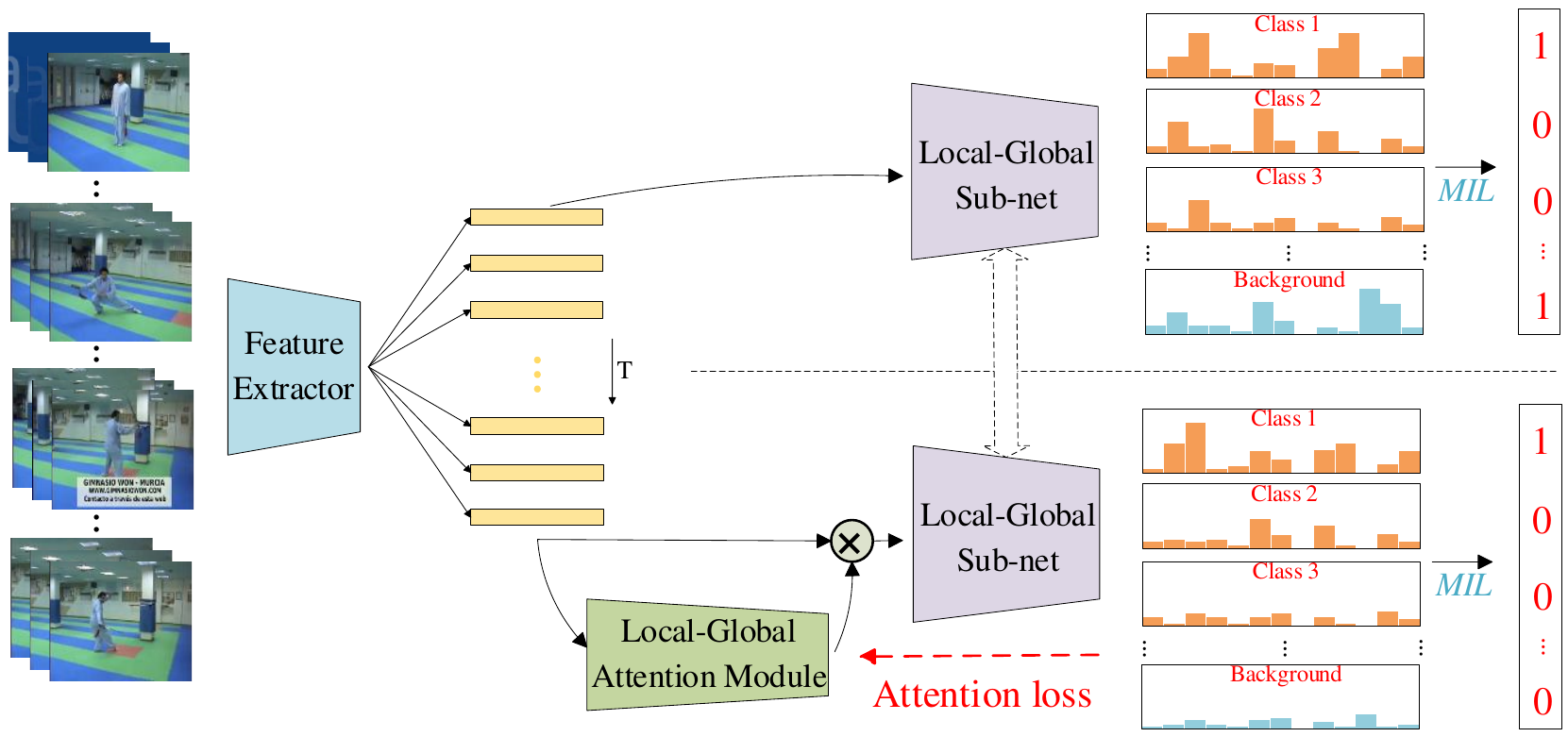}} 
\vspace{-1mm}
\caption{\label{figure-1} 
    The overall architecture of our Local-Global Background Modeling Network (LGBM-Net). 
    Using a pre-trained model, we extract the clip features for the input video, which are then fed into two branches.
    Both branches, sharing local-global sub-net weights, produce class activation sequence (CAS) to predict video-level labels.
    Note that the ground-truth background category of the upper branch is always $1$, and the ground-truth background category of the lower branch is always $0$.
    For some other details, you can refer to BaSNet~\cite{BaSNet}.
}
\vspace{-1mm}
\end{figure*}

%
%
\subsection{Channel-separated convolutional network}
Inspired by the accuracy gains and good computational savings demonstrated by 2D separable convolutions in image classification. Du~\etal~\cite{csn} propose a set of architectures for video classification – 3D Channel-Separated Networks (CSN) – in which all convolutional operations are separated into either pointwise $1\times1\times1$ or depthwise $3\times3\times3$ convolutions. CSN shows that excellent accuracy/computational cost balances can be obtained by leveraging channel separation to reduce FLOPs and parameters as long as high values of channel interaction are retained. Due to its excellent performance in action recognition, we use Kinetics400~\cite{I3D} pre-trained CSN as one of our feature extractors.

\subsection{Slowfast}
Slowfast~\cite{slowfast} model involves two pathways operating at different frame rates. One path-way is designed to capture semantic information that can be given by images or a few sparse frames, and it operates at low frame rates and slow refreshing speed. In contrast, the other pathway is responsible for capturing rapidly changing motion, by operating at fast refreshing speed and high temporal resolution. Note that despite its high temporal rate, this pathway is made very lightweight. This is because this pathway is designed to have fewer channels and weaker ability to process spatial information, while such information can be provided by the first pathway in a less redundant manner. In the competition, We use Slowfast101 pre-trained on Kinetics400 dataset and Slowfast152 pre-trained on Kinetics700~\cite{k700} dataset as backbone.

\subsection{Temporal Difference Network}
Temporal Difference Network (TDN~\cite{TDN}) focuses on capturing multi-scale temporal information for action recognition. The core of TDN is to devise an efficient temporal module by explicitly leveraging a temporal difference operator, and systematically assess its effect on short-term and long-term motion modeling. Meanwhile, TDN is established with a two-level difference modeling paradigm to fully capture temporal information over the entire video. Specifically, for local motion modeling, temporal difference over consecutive frames is used to supply 2D CNNs with finer motion pattern, while temporal difference across segments is incorporated to capture long-range structure for motion feature excitation. TDN provides a simple and principled temporal modeling framework and is selected as our backbone. In the competition, we pre-train TDN on Kinetics700 dataset.

\subsection{ViViT}
Inspired by the large-scale application and good effects of transformer~\cite{transformer} in the field of vision~\cite{ViT}, ViViT~\cite{VIVIT} proposes to use Transformer as basic block to model the relations between temporal and space separately. ViViT is a pure Transformer based model for action recognition. We apply the ViViT-B/16x2 version with factorised encoder, which initialized from imagenet pretrained Vit~\cite{ViT}, and then pre-train it on Kinetics700 dataset. Specifically, we use AdamW as our optimizer and set the base learning rate to 0.0001. The weight decay is set to 0.1. The training is warmed up with 2.5 epochs, with the start learning rate as 1e-6.

\begin{table*}[t]
\small
\centering
\tablestyle{4pt}{1.5}
\begin{tabular}{lcc|ccccccccccc}
& & \multicolumn{12}{c}{mAP@IoU (\%)} \\ \cline{4-13}
\multicolumn{1}{l|}{Method} & Feature & Pre-train & 0.50 & 0.55 & 0.60 & 0.65 & 0.70 & 0.75 & 0.80 & 0.85 & 0.90 & 0.95 & Average mAP \\ 
\shline

\multicolumn{1}{l|}{BaSNet}  & Slowfast101 & K400 & 27.4 & 24.9 & 22.6 & 20.2 & 17.9 & 15.6 & 13.3  & 10.8 & 8.0 & 5.1 & 16.6  \\ 
\multicolumn{1}{l|}{LGBM-Net}  & Slowfast101 & K400 & 29.5 & 27.1 & 24.8 & 22.4 & 20.2 & 17.9 & 15.3 & 12.6 & 9.7 & 6.1 & \bf{18.6} \\ \hline

\multicolumn{1}{l|}{BaSNet}  & Slowfast152 & K700 & 28.2 & 25.5 & 23.2 & 20.9 & 18.6 & 16.4 & 13.8  & 11.1 & 8.4 & 5.1 & 17.1  \\ 
\multicolumn{1}{l|}{LGBM-Net}  & Slowfast152 & K700 & 32.8 & 29.8 & 26.9 & 24.1 & 21.1 & 18.5 & 15.7 & 12.8 & 9.6 & 5.6 & \bf{19.7} \\ \hline

\multicolumn{1}{l|}{BaSNet}  & CSN & K400 & 29.9 & 27.2 & 24.9 & 22.7 & 20.2 & 17.8 & 14.9  & 12.4 & 9.5 & 5.4 & 18.5  \\ 
\multicolumn{1}{l|}{LGBM-Net}  & CSN & K400 & 33.9 & 30.6 & 27.6 & 24.8 & 22.0 & 19.1 & 16.2 & 13.1 & 9.6 & 5.4 & \bf{20.2} \\ \hline

\multicolumn{1}{l|}{BaSNet}  & TDN & K700 & 27.5 & 24.5 & 22.0 & 19.7 & 17.5 & 15.2 & 12.8  & 10.5 & 7.9 & 4.7 & 16.2  \\ 
\multicolumn{1}{l|}{LGBM-Net}  & TDN & K700 & 29.1 & 26.8 & 24.4 & 22.0 & 19.6 & 17.1 & 14.6 & 11.8 & 8.8 & 5.3 & \bf{18.0} \\ \hline

\multicolumn{1}{l|}{BaSNet}  & ViViT & K700 & 27.9 & 25.4 & 23.0 & 20.9 & 18.5 & 16.0 & 13.6  & 10.9 & 8.0 & 4.8 & 16.9  \\ 
\multicolumn{1}{l|}{LGBM-Net}  & ViViT & K700 & 29.2 & 26.8 & 24.4 & 22.2 & 19.7 & 17.6 & 15.2  & 12.8 & 9.9 & 5.9 & \bf{18.4} \\ \hline

\multicolumn{1}{l|}{\textbf{Ensemble}}  & - & - & 37.0 & 33.9 & 30.9 & 28.0 & 24.9 & 22.0 & 18.9  & 15.7 & 12.1 & 7.2 & \bf{23.0} (\textcolor{blue}{\textbf{test: 22.45}}) \\

\end{tabular}
\caption{Performance comparison for different features on validation set of HACS~\cite{HACS} dataset in terms of mAP (\%).}
\label{detection}
\vspace{+3mm}
\end{table*}

\section{Local-Global Background Modeling Network}
In this section, we will introduce the process of our Local-Global Background Modeling Network (LGBM-Net), and then give the experimental results. The architecture of LGBM-Net is showed in Figure~\ref{figure-1}.

\subsection{Local-Global Attention Module}
The goal of local-global attention module is to suppress background frames/segments by the opposite training objective for the background class. Local-global attention module consists of local operation (Conv) and global operation (LSTM~\cite{greff2016lstm}). Note that the two operations are trained in parallel and merged through two convolutional layers followed by sigmoid function. The output of the module is foreground weights which range from $0$ to $1$. At the same time, in order to train the attention of a specific category, we 
use the activation of the highest category in CAS to supervise the attention output in local-global attention module.

\subsection{Local-Global Sub-net}

Local-global sub-net is used to generate CAS, which can be used to predict segment-level class scores. Like local-global attention module, Local-global sub-net also contains local operation (Conv) and global operation (LSTM). Note that we also tried other global operations (\eg, non-local~\cite{nonlocal} and global pooling) for the final ensemble. Afterwards, we aggregate segment-level class scores to derive a video-level class score. Here, we adopt top-k mean technique for training.

\subsection{Detection Results}
After getting the CAS, we can use the watershed algorithm to get the detection result. Note that we only use the CAS of the lower branch in Figure~\ref{figure-1} to generate detection results because of separating foreground and background.

From Table~\ref{detection}, we can draw the following conclusions: (1) It can be seen from the results of Slowfast101 and Slowfast152 that large-scale pre-training and deeper models play a great role in improving performance. (2) The Transformer-based method (ViViT) generally has a slightly worse detection performance than the CNN methods (\eg, CSN and Slowfast). (3) From the results of ensemble, it can be seen that there is complementarity between the models.

\section{Conclusion}

In this report, we propose LGBM-Net, which is based on BaSNet and can well separate the foreground and background. We conduct experiments on multiple features (\eg, CSN, Slowfast, TDN and ViViT) to show the effectiveness of LGBM-Net. Particularly, through ensemble strategy, the detection performance can be further improved.

\section{Acknowledgment}
This work is supported by the National Natural Science Foundation of China under grant (61871435, 61901184), and the Fundamental Research Funds for the Central Universities no.2019kfyXKJC024.

{\small
\bibliographystyle{ieee_fullname}
\bibliography{egbib}
}

\end{document}